\documentclass[runningheads]{llncs}
\usepackage[T1]{fontenc}
\usepackage{graphicx,verbatim}
\usepackage{amsfonts}
\usepackage[table, dvipsnames]{xcolor}
\usepackage{bbm}

\usepackage{algorithm}
\usepackage{algpseudocode}
\usepackage{amsmath}
\usepackage{multirow}
\usepackage{tabularx}
\usepackage{graphicx}
\usepackage{tocbibind}
\begin{document}
%

\title{Ambiguous Medical Image Segmentation \\ Using Diffusion Schr\"{o}dinger Bridge}
%
\author{Lalith Bharadwaj Baru*$^1$, Kamalaker Dadi$^1$, \\ Tapabrata Chakraborti*$^2$, Raju S. Bapi$^1$}
\authorrunning{Lalith Bharadwaj Baru et al.}
%
\institute{$^1$International Institute of Information Technology, Hyderabad, India.\\ 
$^2$The Alan Turing Institute and University College London, London, UK.\\
\email{*lalith.baru@research.iiit.ac.in, tchakraborty@turing.ac.uk}}



\maketitle              
\begin{abstract}

Accurate segmentation of medical images is challenging due to unclear lesion boundaries and mask variability. We introduce \emph{Segmentation Sch\"{o}dinger Bridge (SSB)}, the first application of Sch\"{o}dinger Bridge for ambiguous medical image segmentation, modelling joint image-mask dynamics to enhance performance. SSB preserves structural integrity, delineates unclear boundaries without additional guidance, and maintains diversity using a novel loss function. We further propose the \emph{Diversity Divergence Index} ($D_{DDI}$) to quantify inter-rater variability, capturing both diversity and consensus. SSB achieves state-of-the-art performance on LIDC-IDRI, COCA, and RACER (in-house) datasets.

\keywords{Ambiguous Medical Image Segmentation  \and Sch\"{o}dinger Bridge \and Diffusion Models $\; \;$  \and Generative Modeling $\;$ \and Uncertainty Quantification}

\end{abstract}

\section{Introduction}\label{sec-intro}

Diagnosis, a cornerstone of medicine, heavily depends on individual assessment strategies. Recent studies highlight the prevalence of misdiagnosis even for common health conditions \cite{kurvers2016boosting}, emphasizing the importance of reducing diagnostic errors. Medical image segmentation, crucial in artificial intelligence (AI) enabled clinical decision support systems (CDSS), often relies on deterministic deep learning models that predict only a single mask per image \cite{ronneberger2015u}. However, varying clinician opinions on anomalies \cite{alpert2004quality} result in low diagnostic consensus. Incorporating multiple expert interpretations improves diagnosis and reduces false negatives, but clinician time is a costly resource. Deterministic models have advanced medical image segmentation but often produce suboptimal results due to their bias toward the most likely hypothesis, while pixel-wise uncertainty models can yield inconsistent outputs. This necessitates models that capture diverse expert interpretations, such as the Probabilistic U-Net \cite{kohl2018probabilistic}, PhiSeg-Net \cite{baumgartner2019phiseg},
and CIDM \cite{rahman2023ambiguous} etc.

Previous work in ambiguous segmentation has explored variational autoencoders (VAEs) and Bayesian methods to generate diverse segmentation masks. Kohl \emph{et al.} \cite{kohl2018probabilistic} introduced the Probabilistic U-Net, combining a U-Net with a conditional VAE for multiple hypothesis generation. While effective, it struggles with mask diversity and requires significant computational resources. PhiSeg-Net \cite{baumgartner2019phiseg} enhanced diversity using a hierarchical VAE with a U-Net backbone, but its axis-aligned Gaussian latent space \cite{selvan2020uncertainty} remains restrictive, limiting expressiveness in modeling uncertainties. Rahman \emph{et al.} \cite{rahman2023ambiguous} and Zbinden \emph{et al.} \cite{zbinden2023stochastic} both leverage diffusion models based on Gaussian distributions. A core challenge in ambiguous segmentation is capturing structural lesion features and generating plausible boundaries. Current diffusion models sample from Gaussian noise, making it difficult to represent fine-grained lesion structures. Although using input images as guidance improves CT image representations, it often fails to preserve detailed lesion characteristics critical for precise segmentation. In this work we address some of these drawbacks through the following contributions: \\
    - We empirically demonstrate the effectiveness of Sch\"{o}dinger Bridge (SB) in ambiguous medical image segmentation by introducing Segmentation Sch\"{o}dinger Bridge (SSB), which adapts the SB framework and outperforms existing benchmarks on LIDC-IDRI, Stanford COCA, and our in-house RACER datasets.\\
    - To address ambiguity in segmentation, we propose a novel loss function designed to guide the model in handling diverse variations among segmentation masks. Additionally, we introduce a new metric, the \textit{Diversity Divergence Index} (\(D_{DDI}\)), which quantifies inter-annotator variability and agreement, capturing both diversity and consensus among expert annotations.\\
    - Our approach delivers significant improvements across key metrics (\(GED\), \(D_{max}\), \(\text{CI Score}\), and \(D_{DDI}\)) compared to existing baselines. Moreover, it showcases robustness, offering fast sampling rates and reliable segmentation outputs suitable for real-world applications.

\section{Datasets and Setup}
\paragraph{\bf Datasets:} We use the publicly available LIDC-IDRI dataset \cite{armato2004lung}, comprising of lung CT scans with lesion segmentations by four radiologists (1084 subjects, train:test split = 9:1). Additionally, we utilize the Stanford COCA dataset (787 subjects, train:test split = 3:1) with one expert annotations. We also test our model an in-house dataset RACER with data from 44 subjects and annotations from 2 experts with demographic and ethnic variations distinct from COCA, enabling a robust evaluation of model generalization. The evaluation follows the protocol outlined by Rahman \emph{et al.} \cite{rahman2023ambiguous}.


\paragraph{\bf Evaluation Metrics:} 
The Generalized Energy Distance (GED) is a statistical metric widely used to evaluate generative models in ambiguous medical image segmentation \cite{szekely2013energy}, \cite{kohl2018probabilistic}, \cite{baumgartner2019phiseg}, \cite{rahman2023ambiguous}. 
However, Rahman \emph{et al.} \cite{rahman2023ambiguous} identified shortcomings in GED for capturing both segmentation performance and diversity, introducing complementary metrics such as Collective Intelligence Score (CI Score), Maximum Dice Matching (\(D_{max}\)), and Diversity Agreement (\(D_{\mathcal{A}}\)). In this work, we employ GED alongside these metrics for a rigorous and holistic evaluation while highlighting the limitations of \(D_{\mathcal{A}}\). To address this, we propose the \(D_{DDI}\), which provides a more robust quantification of diversity.

\section{Methods}\label{sec-meth}
\subsection{Score-based Generative Models and Sch\"{o}dinger Bridge}
Before introducing Score-based Generative Models (SGMs) and the Sch\"{o}dinger Bridge (SB), we establish standard notations. Let \( X_t \in \mathbb{R}^d \) be a stochastic process over continuous time \( t \sim \mathcal{U}[0,1] \), with initial (\textit{data}) distribution \( p_A \) and terminal (\textit{mask}) distribution \( p_B \). The Wiener process and its reverse, based on Anderson et al. \cite{anderson1982reverse}, are \( W_t \) and \( \overline{W}_t \) in \( \mathbb{R}^d \), while \( \mathbb{I} \in \mathbb{R}^{d \times d} \) is the identity matrix. The reference path measure \( \mathbb{P} \) corresponds to the forward SDE dynamics, serving as a baseline for optimization, while the target path measure \( \mathbb{Q} \) represents the optimal transport process constrained by \( p_A \) (at \( t = 0 \)) and \( p_B \) (at \( t = 1 \)). The functions \( \Psi(t, x) \) and \( \widehat{\Psi}(t, x) \) solve the forward and backward PDEs, respectively, defining the drift components that govern SB dynamics.

\paragraph{\bf Score-based Generative Models (SGMs)} aim to perturb data across continuous noise scales using stochastic differential equations (SDEs) and reconstruct the data distribution via reverse-time SDEs. The reverse process, guided by the score function, enables transformations between arbitrary distributions and Gaussians \cite{song2020score}. Given data \( X_0 \sim p_A \), the forward SDE perturbs it, while the reverse SDE reconstructs it using learned scores. Here, the drift and diffusion coefficients govern the noise and transformation. SGMs employ a U-Net \cite{ronneberger2015u} to parameterize the score function, optimized via a denoising score-matching loss \cite{vincent2011connection}, ensuring sampled distributions closely resemble \( p_A \).
\begin{align}\label{eq-allsgms}
    & \text{Forward SDE:} \quad dX_t = f_t(X_t) \, dt + \sqrt{\beta_t} \, dW_t, \quad X_0 \sim p_{A}, \nonumber \\
    & \text{Reverse SDE:} \quad dX_t = [f_t - \beta_t \nabla \log p(X_t, t)] \, dt + \sqrt{\beta_t} \, d\overline{W}_t, \nonumber \\
    & \text{Training Loss:} \quad \mathbb{E}\big[ \lambda(t) \|\epsilon_\theta(X_t, t) - \sigma_t \nabla \log p(X_t, t \mid X_0)\|^2 \big], \nonumber \\
    & \text{Sampling SDE:} \quad dX_t = \left[f - \beta_t \epsilon_\theta(t, X_t)\right] dt + \sqrt{\beta_t} dW_t, \quad X_1 \sim p_B.
\end{align} 

\begin{figure}[!t]
    \centering
    \includegraphics[width=0.99\textwidth]{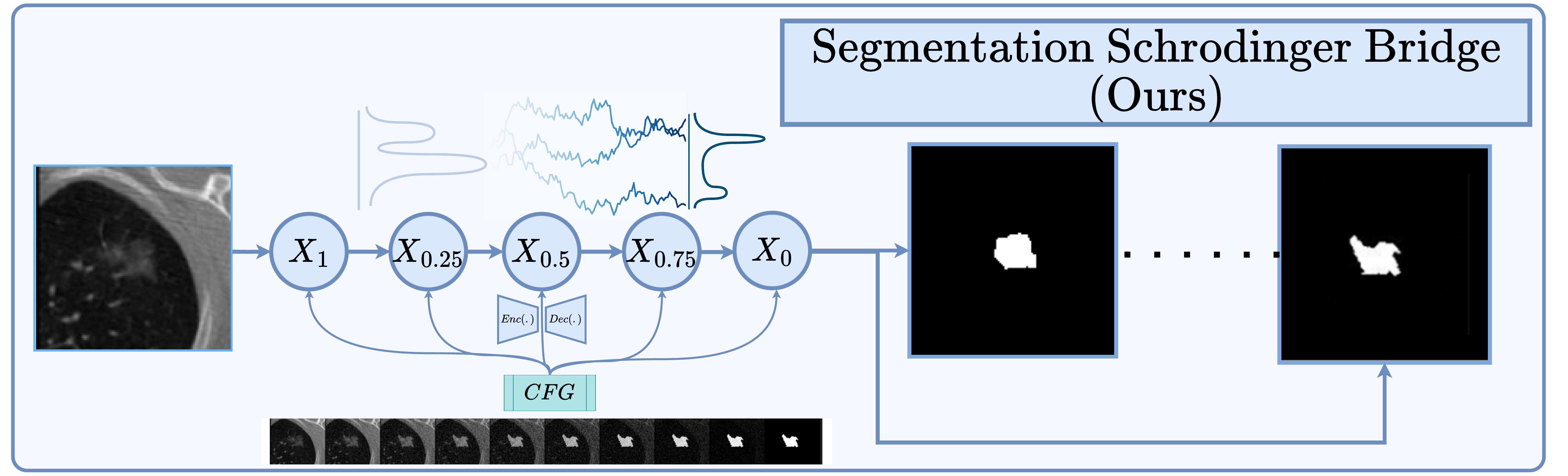}
    \caption{\textbf{Overview of the Segmentation Schrödinger Bridge (SSB) Framework:} The model formulates segmentation as a stochastic transport problem and progressively maps the input image distribution to the segmentation mask distribution while preserving structural integrity. Classifier-Free Guidance (CFG) improves control over diversity and enables the generation of expert-aligned yet varied segmentations. Time-dependent U-Net parameter optimization and refined residual connections enhance performance.}
    \label{fig-ssb-example}
\end{figure}

Here, \( \nabla \log p(X_t, t \mid X_0) \) is computed analytically, with \( \sigma_t^2 \) scaling the regression target and \( \lambda(t) \) acting as a tunable hyperparameter influencing performance \cite{ho2020denoising}. This allows SGMs to transform \( p_A \) into \( p_B \) through learned scores. However, SGMs are limited in flexibility, as they primarily transition between image distributions and Gaussian noise, restricting efficient mapping to arbitrary target distributions. To overcome this, we explore the Sch\"{o}dinger Bridge approach, which offers greater adaptability with minimal data manipulation.

\paragraph{\bf SGMs and Schr\"{o}dinger Bride (SB)}
The SB framework minimizes the KL divergence between a reference path measure \( \mathbb{P} \) and a target path measure \( \mathbb{Q} \), subject to prescribed marginal densities \( p_A \) (at \( t = 0 \)) and \( p_B \) (at \( t = 1 \)) \cite{pavon1991free}. The optimization is formulated as \(
\min_{\mathbb{Q} \in P(p_{A}, p_{B})} D_{KL}(\mathbb{Q} \, || \, \mathbb{P}), \) where \( \mathbb{P} \) is typically chosen as the path measure of the forward SDE. The SB framework, as described in \cite{chen2021stochastic}, \cite{chen2021likelihood}, optimally transports \( p_A \) to \( p_B \) by solving PDEs for forward and backward drifts. Unlike SGMs, SB introduces non-linear components, making it more adaptable to prior data variations. While SB’s stochastic processes resemble SGMs, they incorporate additional coupling terms, resulting in a factorized marginal density. The governing equations are,
\begin{align}\label{eq-sb-summary}
    & \text{SB Forward SDE:} \quad dX_t = [f_t + \beta_t \nabla \log {\Psi}(X_t, t)] \, dt + \sqrt{\beta_t} \, d{W}_t, \nonumber \\
    & \text{SB Backward SDE:} \quad dX_t = [f_t - \beta_t \nabla \log \widehat{\Psi}(X_t, t)] \, dt + \sqrt{\beta_t} \, d\overline{W}_t,  
\end{align}
As demonstrated in \cite{pavon1991free}, \cite{chen2021stochastic}, \cite{chen2021likelihood} this framework extends SGMs, with forward-backward SDEs that resemble those in SGMs but include non-linear drifts (\( \nabla \log \Psi \) and \( \nabla \log \widehat{\Psi} \)) (equation \ref{eq-sb-summary}) to handle diverse prior data effectively \cite{anderson1982reverse}, \cite{song2020score}.

\paragraph{\bf Optimal Boundary Constraints}
To address the computational challenges of nonlinear SBs, linear SDEs are formulated by redesigning the forward and backward drifts to solve the Fokker-Planck equation \cite{risken1996fokker}, \cite{liu2023i2sb}. These linear SDEs simplify SB coupling constraints, improving computational efficiency and tractability. While SBs nonlinear drifts correspond to score functions, sampling requires parameterizing \( \nabla \log \widehat{\Psi} \) via a score network. Dirac delta distributions are applied to further streamline boundary conditions, ensuring efficient convergence to a target state \( \nu \). The resulting equations are formulated as follows,
\begin{align}\label{eq-linear-sbs}
    & \text{Linear Forward SDE:} \quad dX_t = f_t(X_t) \, dt + \sqrt{\beta_t} \, dW_t, \quad X_0 \sim \Psi(\cdot, 0), \nonumber \\
    & \text{Linear Backward SDE:} \quad dX_t = f_t(X_t) \, dt + \sqrt{\beta_t} \, d\overline{W}_t, \quad X_1 \sim \widehat{\Psi}(\cdot, 1), \\
    & \text{Boundary Condition:} \quad p_A(\cdot) = \delta_\nu(\cdot), \quad p_B = \Psi(\cdot, 1)\widehat{\Psi}(\cdot, 1).\nonumber 
\end{align}
These linear SDEs ensure that the reverse process converges to the Dirac delta \( \delta_\nu(\cdot) \), allowing efficient computation of the score function \( \nabla \log p(X_t, t \mid X_0 = \nu) \) for each sample \( \nu \). This reformulation balances mathematical rigor with computational efficiency, making SB scalable and practical for real-world applications.

\subsection{Segmentation Schr\"{o}dinger Bridge (SSB)}
Theoretical insights necessitate learning \( \nabla \log \widehat{\Psi}(\cdot) \) in a tractable manner. We apply this to ambiguous medical image segmentation by training a diffusion model that directly maps a given CT image to the lesion. To fully describe the diffusion process, we follow a structured design strategy \cite{chang2023design}, detailing the training, generation (sampling), and objective function.

 
\paragraph{\bf Training and Generation:} Preserving region-of-interest (ROI) boundaries is crucial in segmentation. Traditional forward diffusion degrades structural image properties, reducing mask quality. Prior methods like CIMD \cite{rahman2023ambiguous} and CCDM \cite{zbinden2023stochastic} generate masks by diffusing from Gaussian noise with image feature guidance, but often fail to retain structural lesion details. Inspired by Palette \cite{saharia2022palette}, we introduce a controlled degradation technique that perturbs the image partially, preserving structural integrity. Hence, at each time when controlled noise is injected, the model traverses different random paths and yields unique variations in the mask, even though they converge to the same marginals. Further information about generation and training is provided in Algorithm (\ref{alg-generation}), (\ref{alg-training}). Training scalable diffusion models require efficient computation of \( X_t \), which is intractable using Equation (\ref{eq-sb-summary}) due to nonlinear forward drift and the inability of linear SDEs (\ref{eq-linear-sbs}) to model high-probability regions effectively. To overcome this, we adopt an analytical posterior formulation from Liu et al. \cite{liu2023i2sb}: 

\begin{equation}\label{eq-ssb-posterior}
q(X_t \mid X_0, X_1) = \mathcal{N}\left(X_t; \mu_t(X_0, X_1), \Sigma_t\right)
\end{equation}
Where, \( \mu_t = \frac{\bar{\sigma}_t^2}{\bar{\sigma}_t^2 + \sigma_t^2} X_0 + \frac{\sigma_t^2}{\bar{\sigma}_t^2 + \sigma_t^2} X_1, \quad \Sigma_t = \frac{\sigma_t^2 \bar{\sigma}_t^2}{\bar{\sigma}_t^2 + \sigma_t^2} \cdot \mathbb{I}.\) This allows direct sampling of \( X_t \) during training from \( p_A(X_0) \) and \( p_B(X_1 \mid X_0) \) without solving nonlinear diffusion as in prior Schrödinger Bridge (SB) models \cite{vargas2021solving}. During generation, given only \( X_1 \sim p_B \), running a standard diffusion model \cite{song2020score}, \cite{ho2020denoising} from \( X_1 \) recovers the marginal density of SB paths, provided \( X_{\epsilon, 0} \) approximates \( X_0 \) well. Thus, the sampling method in Equation (\ref{eq-ssb-posterior}) is both tractable and effective in covering the required regions.

\scalebox{0.80}{
\begin{minipage}[!t]{0.55\textwidth}
\begin{algorithm}[H]
\caption{Generation of SSB Model}\label{alg-generation}
\begin{algorithmic}[1]
\State \textbf{Input:} Trained network $\epsilon_\theta$, noise schedule $\{\beta_t\}$
\State \textbf{Output:} Generated samples $X_0$
\Procedure{Generation}{$\epsilon_\theta$, $\{\beta_t\}$}
    \State $X_N \sim p_B(X_N)$
    \For{$t = N, \ldots, 1$}
        \State Predict Mask: $\hat{X}_0$ via $\epsilon_\theta (X_t,l, t)$
        \State  Sample Masks \cite{ho2020denoising}: \\ $\quad \quad \;\;$ $X_{t-1} \sim p(X_{t-1}|\hat{X}_0, X_t)$
    \EndFor
    \State \textbf{Return} $X_0$
\EndProcedure
\end{algorithmic}
\end{algorithm}
\end{minipage}
\hfill\hspace{4pt}
\begin{minipage}[!t]{0.60\textwidth}
\begin{algorithm}[H]
\caption{Training of SSB Model}\label{alg-training}
\begin{algorithmic}[1]
\State \textbf{Input:} Training data $\mathcal{D}$; U-Net $\epsilon_\theta$, Experts $\eta$
\State \textbf{Output:} Trained network $\epsilon_\theta$
\Procedure{Training}{$\mathcal{D}$, $\epsilon_\theta$}
    \While{not converged}
        \State $t,l \sim \mathcal{U}[0,1], \mathcal{U}[1, \eta]$
        \State $X_0 \sim p_A(X_0)$, $X_1 \sim p_B(X_1|X_0)$
        \State  $X_t \sim q(X_t|X_0, X_1)$ (ref eq. (\ref{eq-ssb-posterior}))
        \State Optimize \cite{liu2023i2sb}: $\mathcal{L}_{SSB}$ ($\epsilon_\theta,l,t$) 
        \State Update $\epsilon_\theta$ with GD to $\min$ $\mathcal{L}_{SSB}$.
    \EndWhile
    \State \textbf{Return} Trained $\epsilon_\theta$
\EndProcedure
\end{algorithmic}
\end{algorithm}
\end{minipage}
}

\paragraph{\bf Loss Function:}
Our loss function differs from the standard ADM approach \cite{dhariwal2021diffusion} by leveraging Classifier-Free Guidance (CFG) \cite{ho2022classifier}. As outlined in Algorithm \ref{alg-training}, we first determine the number of experts and stochastically sample expert labels during training. The model then estimates noise in both conditional and unconditional settings: i) \textit{Conditional}: \( \epsilon_\theta(x_t, y, t) \) using data-label pairs \( (x_0, y) \); ii) \textit{Unconditional:} \( \epsilon_\theta(x_t, t) \) without label. During sampling, noise is estimated via linear interpolation \( \hat{\epsilon} = (1 + \omega)\epsilon_\theta(x_t, y, t) -\omega\epsilon_\theta(x_t, t), \) enabling CFG-based reconstruction of \( x_{t-1} \). This integration ensures diverse yet structurally consistent segmentations with fine-grained delineation.  A key distinction lies in computing the score function in the loss, formulated for the reverse linear drift in Equation (\ref{eq-linear-sbs}) while incorporating CFG. Given the segmentation focus, we further enhance precision by integrating a Dice loss, emphasizing fine boundary details to improve segmentation accuracy. The complete loss formulation is provided below,
\begin{equation}\label{eq-ssbloss}
    \mathcal{L}_{\text{SSB}} = \mathbb{E} \left[\left| \left|\epsilon_\theta(X_t,l, t) - \frac{(X_t - X_0 )}{\sigma_t}  \right|\right|^2  - 2 \gamma \cdot \frac{||X_0 \cap \epsilon_\theta(X_t,l, t)||}{||X_0|| + ||\epsilon_\theta(X_t,l, t)||} + \gamma  \right].
\end{equation}

\subsection{Diversity Divergence Index}
In ambiguous medical image segmentation, a critical objective is to generate diverse expert-like predictions while capturing inter-annotator variability and agreement. To quantify both diversity and consensus among expert annotations, we propose the \textit{Diversity Divergence Index} (\(D_{DDI}\)). The metric is formulated as follows. Given a model that generates \(N\) segmentation masks for an image and \(M\) expert annotations, we compute the \texttt{Dice score} for each generated mask against each expert, forming an \(M \times N\) matrix. Each row, denoted as \textit{expert distribution} \(\text{exp-dist}_{i} \in \mathbb{R}^M\), represents the Dice scores between all generated masks and the \(i^{\text{th}}\) expert. Similarly, each column, referred to as the \textit{generated distribution} (\(\text{gen-dist}_{j} \in \mathbb{R}^M\)), captures the Dice scores between a single generated mask and all expert annotations, reflecting how closely the model's predictions align with expert variability.

Next, we compute the Jensen-Shannon (JS) divergence \cite{lin1991divergence} across expert distributions, measuring the divergence between expert predictions. Finally, \(D_{DDI}^\text{exp}\) aggregates these pairwise divergences, providing a comprehensive measure of inter-expert variability and \(D_{DDI}^\text{gen}\) aggregates these pairwise divergences, providing a comprehensive measure of inter-mask variability : 

\begin{align}
    D_{DDI}^\text{exp} = & \frac{3}{^MC_2}\sum_i \sum_j \mathbbm{1}_{[i\neq j]} JS (\text{exp-dist}_{i}, \text{exp-dist}_{j})\\
    D_{DDI}^\text{gen} = & \frac{3}{^NC_2}\sum_i \sum_j \mathbbm{1}_{[i\neq j]} JS (\text{gen-dist}_{i}, \text{gen-dist}_{j})
\end{align}

Here, $\mathbbm{1}_{[i\neq j]}$ is the indicator function, which excludes the same expert comparison. This formulation captures both \textit{diversity among expert masks} and \textit{agreement in generated masks}, making it a robust metric for evaluating model uncertainty in ambiguous medical image segmentation.

\begin{figure}[!t]
    \centering
    \includegraphics[width=0.999\textwidth]{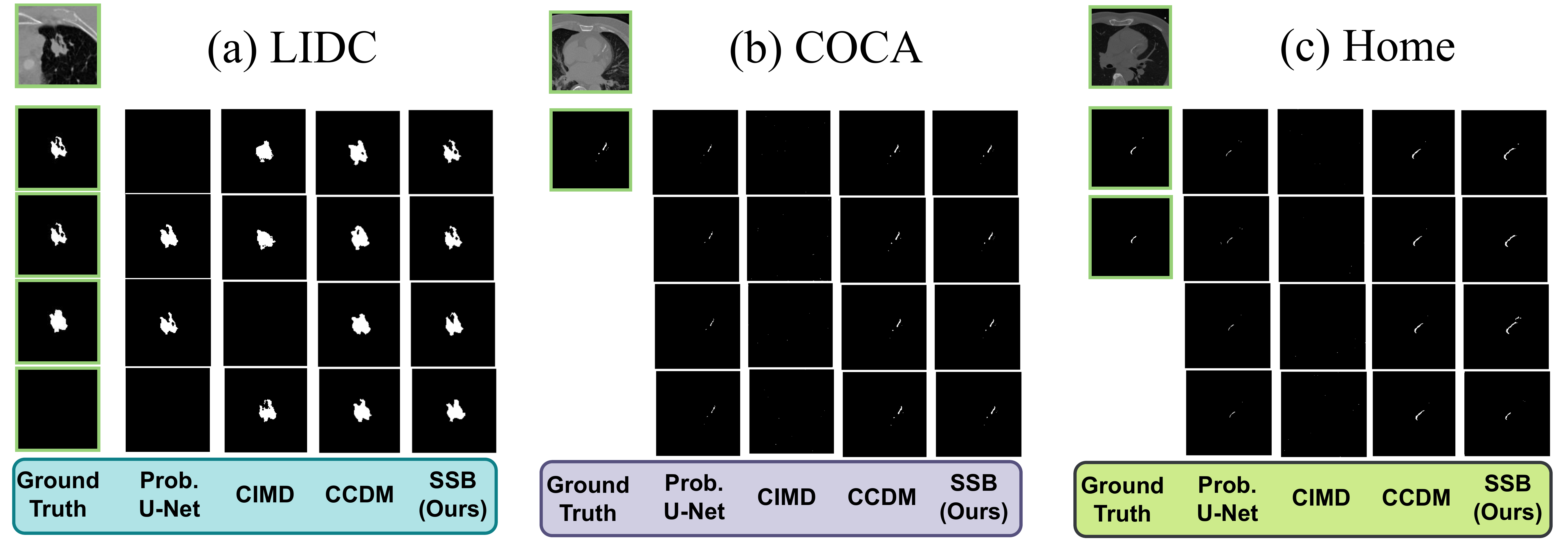}
    \caption{The first image Fig. 2(a) from LIDC-IDRI \cite{armato2004lung} shows that Probabilistic U-Net fails to generate accurate masks, while CIMD aligns with expert annotations but lacks precise lesion boundaries. CCDM performs suboptimal, whereas SSB achieves fine-grained boundary delineation while maintaining segmentation diversity. In Fig. 2(b) from Stanford COCA \cite{eng2021automated}, CIMD fails to capture calcium plaques, and while Prob. U-Net and CCDM miss very small plaques, and SSB successfully segments all deposits. For the third image (c) from RACER (in-house), trained only on COCA, SSB generalizes well, producing diverse yet expert-aligned annotations, unlike other methods that fail to overlap with experts or provide meaningful variability. In  Fig. 2(c) the final output closely matches the second expert, while the remaining three exhibit controlled diversity}
    \label{fig-ssb-viz}
\end{figure}

\section{Results and Discussion}
Table~\ref{tab-main} presents a comparative analysis of our proposed SSB variants against state-of-the-art baseline methods across the LIDC-IDRI, COCA, and Home datasets. SSB++ extends the capabilities of SSB by incorporating Classifier-Free Guidance (CFG), enhancing control over the generative process to achieve an optimal balance between fidelity and diversity in segmentation. While SSB adopts the standard parametric configuration of DDPM~\cite{ho2020denoising}, SSB++ introduces task-specific optimizations to time-dependent U-Net parameters, including the refinement of up-sampling and down-sampling layers and residual connections to enhance feature extraction and reconstruction. These architectural advancements make SSB++ a more robust and efficient solution for ambiguous medical image segmentation, consistently delivering superior performance.\footnote{Complete training details of SSB++, including optimized parametric configurations, will be made available in the code repository.} Notably, SSB++ consistently surpasses prior approaches across all key metrics, achieving substantial improvements over the strongest baselines. The core limitation of existing diffusion methods such as CIMD \cite{rahman2023ambiguous} and CCDM \cite{zbinden2023stochastic} lies in their generative process itself, which starts from Gaussian noise and progressively refines segmentations. This often leads to a loss of structural integrity, resulting in coarse or anatomically inconsistent predictions. In contrast, our approach begins from the input image itself and progressively transforms it into the segmentation mask, ensuring that the anatomical structure is well-preserved throughout the process. Among prior works, CCDM performs closely yet it still struggles to maintain structural fidelity compared to ours. Across all datasets, SSB++ achieves significant performance improvements over prior methods. For instance, on the LIDC-IDRI dataset, SSB++ reduces GED by 21.2\% compared to CCDM, while simultaneously improving $D_{max}$ by 5.7\% and CI by 4.4\%, indicating that our method generates segmentations that are both structurally accurate and diverse. Similar trends are observed in COCA, where SSB++ achieves a 25.6\% lower GED than CIMD, and on the in-house RACER dataset, where it further improves GED by 19.3\% and CI by 5.3\% over CCDM. Qualitative results are presented in Fig. 2.

\begin{table*}[t!]
\caption{The results were obtained from the test set of the LIDC-IDRI, COCA, and RACER datasets. Results for PhiSeg \cite{baumgartner2019phiseg} 
was taken from their original research (denoted $^\dag$), while CIMD \cite{rahman2023ambiguous} and CCDM \cite{zbinden2023stochastic} were fully reproduced. Metrics with ($\downarrow$) favor lower values, while ($\uparrow$) indicate higher is better. Also, $D_{DDI}$ entries for the COCA dataset are empty as there are single expert masks. The reported results on the test set were statistically significant (p < 0.05).
}\label{tab-main}
\scalebox{0.60}{
\begin{tabular}{cc}
\begin{minipage}{0.48\linewidth}
\begin{tabular}{lccccccc}
\hline
\rowcolor{lightgray}
\multicolumn{8}{c}{\textbf{LIDC-IDRI Dataset}} \\ 
\textbf{Method} & \textbf{NFE} & \textbf{$GED$} $\downarrow$ 
& \textbf{$D_{max}$} $\uparrow$ & $CI$  $\uparrow$ & \textbf{$D_\mathcal{A}$} $\uparrow$ &  $D_{DDI}^\text{exp}$ $\uparrow$ & $D_{DDI}^\text{gen}$ $\uparrow$\\ 
\hline \hline
\rowcolor{blue!20}
Prob. UNet  \cite{kohl2018probabilistic}     & -            & 0.310  
& 0.907              & 0.831                & 97.9  & 0.391 &0.287 \\
\rowcolor{blue!20}
PhiSeg$^\dag$ \cite{baumgartner2019phiseg}  & -           
& 0.270            &    0.904            & 0.736                 & -  & - & -  \\
\hline
\rowcolor{green!20}
\rowcolor{red!20}
CIMD    \cite{rahman2023ambiguous}         & 1000         & 0.318           
& 0.841           & 0.883            & 0.996   & 0.341 & 0.214 \\
\rowcolor{red!20}
CCDM    \cite{zbinden2023stochastic}         & 1000         & 0.264            
& 0.886              & 0.918               & 0.997  & 0.402 & 0.324 \\ 
\hline
\rowcolor{violet!20}
SSB (Ours)       & 50           & 0.245               
&  0.925               & 0.945                & 0.995 & 0.397 & 0.317 \\
\rowcolor{violet!20}
SSB++ (Ours)     & 50           &\bf 0.208           
&\bf  0.943           &\bf  0.958          &\bf  0.997 & \bf 0.427 &\bf  0.356 \\
\hline
\end{tabular}
\end{minipage}
\hspace{115pt}
\begin{minipage}{0.48\linewidth}
\begin{tabular}{lccccccc}
\hline
\rowcolor{lightgray}
\multicolumn{8}{c}{\textbf{Stanford COCA Dataset}} \\ 
\textbf{Method} & \textbf{NFE} & \textbf{$GED$} $\downarrow$ 
& \textbf{$D_{max}$} $\uparrow$ & $CI$  $\uparrow$ & \textbf{$D_\mathcal{A}$} $\uparrow$ &  $D_{DDI}^\text{exp}$ $\uparrow$ & $D_{DDI}^\text{gen}$ $\uparrow$\\ 
\hline \hline
\rowcolor{blue!20}
Prob. UNet  \cite{kohl2018probabilistic}     & -            & 0.658  
& 0.722              & 0.789                & 0.999   & - & - \\
\hline
\rowcolor{red!20}
CIMD    \cite{rahman2023ambiguous}         & 1000         & 0.625           
& 0.501           & 0.599            & 0.999  & -   & - \\
\rowcolor{red!20}
CCDM    \cite{zbinden2023stochastic}         & 1000         & 0.5            
& -               & -                & -  & - & - \\
\hline
\rowcolor{violet!20}
SSB (Ours)       & 50           & 0.421               
&  0.891               & 0.908                & 0.999 & - & - \\
\rowcolor{violet!20}
SSB++ (Ours)     & 50           &\bf 0.372           
&\bf  0.941           &\bf  0.955         &  0.999 & - & - \\ 
\hline
\hline
\rowcolor{lightgray}
\multicolumn{8}{c}{\textbf{RACER (Home) Dataset}} \\ 
\rowcolor{blue!20}
Prob. UNet  \cite{kohl2018probabilistic}     & -            & 0.792  
& 0.610              & 0.697                & 0.999   & $\approx$0 &  $\approx$0 \\
\hline
\rowcolor{red!20}
CIMD    \cite{rahman2023ambiguous}         & 1000         & 0.758           
& 0.503           & 0.601            & 0.999  &  $\approx$0 &  $\approx$0   \\
\rowcolor{red!20}
CCDM    \cite{zbinden2023stochastic}         & 1000         & 0.551            
& 0.630              & 0.715                &0.999  & 0.009 & 0.002\\ 
\hline
\rowcolor{violet!20}
SSB (Ours)       & 50           & 0.528               
&  0.635               & 0.712                & 0.999 & 0.008 & 0.002 \\
\rowcolor{violet!20}
SSB++ (Ours)     & 50           &\bf 0.635           
&\bf  0.675           &\bf  0.753          &  0.999 & \bf 0.013 & \bf 0.007\\ 
\hline

\end{tabular}
\end{minipage}
\end{tabular}
}
\end{table*}

A key contribution of this work is the introduction of $D_{DDI}$, a robust metric designed to effectively quantify inter-rater variability and consensus in ambiguous segmentation tasks. Unlike the existing $D_\mathcal{A}$ metric, which remains consistently close to one and fails to meaningfully differentiate diversity across methods, $D_{DDI}$ provides a bounded and interpretable measure of segmentation variability, making it a more reliable indicator of diversity in medical image segmentation. To comprehensively assess diversity agreement, we evaluate both $D_{DDI}^{exp}$, which captures alignment with expert annotations, and $D_{DDI}^{gen}$, which reflects the diversity in generated segmentations. Our proposed SSB++ framework demonstrates a substantial improvement over previous methods, achieving the highest $D_{DDI}^{exp}$ across multiple datasets. On the LIDC-IDRI dataset, SSB++ improves expert agreement diversity by approximately 6.2\% compared to CCDM, while on the in-house RACER dataset, it increases by over 44.4\%. Similarly, in terms of generative diversity, SSB++ surpasses CCDM by 9.8\% on LIDC-IDRI and exhibits a remarkable threefold improvement on the RACER dataset, signifying its ability to generate diverse yet clinically meaningful segmentation masks. Beyond its advancements in diversity modeling, SSB++ also delivers significant computational efficiency. While previous methods such as CIMD and CCDM require 1000 no. of function evaluations (NFEs), SSB++ achieves superior segmentation performance with only 50 NFEs, reducing computational overhead by an order of magnitude. This efficiency makes SSB++ not only a more accurate solution but also a more viable choice for real-world clinical deployment where computational resources are often limited.

\section{Conclusion}
In this work, we present Segmentation Schrödinger Bridge (SSB), a novel framework that eliminates the need of traditional diffusion training or sampling from pure Gaussian noise and thereby by preserves the structural integrity of lesion features. To enhance robustness, we introduce a loss function tailored for segmentation (\(\mathcal{L}_{SSB}\)), alongside a new metric (\(D_{DDI}\)) that effectively quantifies inter-rater variability and agreement. Our approach achieves state-of-the-art performance, surpassing competing methods in both ambiguous segmentation accuracy as well as sampling efficiency across several benchmark datasets on medical image segmentation tasks.

\section*{Acknowledgments}
TC is supported by the Turing-Roche Strategic Partnership and a NIHR Principal Research Fellowship from the UCL Biomedical Research Centre. We acknowledge IHub-Data, IIIT Hyderabad (H1-002), for financial assistance.

\section*{Disclosure of Interest}
The authors declare that they have no financial or non-financial competing interests that could have influenced the work reported in this paper.

\vspace{18pt}
\begingroup
  \let\clearpage\relax      
  \bibliographystyle{plain} 
  \bibliography{Paper-4859}     

\begin{thebibliography}{10}

\bibitem{alpert2004quality}
Hillel~R Alpert and Bruce~J Hillman.
\newblock Quality and variability in diagnostic radiology.
\newblock {\em Journal of the American College of Radiology}, 1(2):127--132, 2004.

\bibitem{anderson1982reverse}
Brian~DO Anderson.
\newblock Reverse-time diffusion equation models.
\newblock {\em Stochastic Processes and their Applications}, 12(3):313--326, 1982.

\bibitem{armato2004lung}
Samuel~G Armato~III et~al.
\newblock Lung image database consortium: developing a resource for the medical imaging research community.
\newblock {\em Radiology}, 232(3):739--748, 2004.

\bibitem{baumgartner2019phiseg}
Christian~F Baumgartner, Kerem~C Tezcan, Krishna Chaitanya, Andreas~M H{\"o}tker, Urs~J Muehlematter, Khoschy Schawkat, Anton~S Becker, Olivio Donati, and Ender Konukoglu.
\newblock Phiseg: Capturing uncertainty in medical image segmentation.
\newblock In {\em Medical Image Computing and Computer Assisted Intervention--MICCAI}, pages 119--127. Springer, 2019.

\bibitem{chang2023design}
Ziyi Chang, George~A Koulieris, and Hubert~PH Shum.
\newblock On the design fundamentals of diffusion models: A survey.
\newblock {\em arXiv preprint arXiv:2306.04542}, 2023.

\bibitem{chen2021likelihood}
Tianrong Chen, Guan-Horng Liu, and Evangelos~A Theodorou.
\newblock Likelihood training of schr{\"o}dinger bridge using forward-backward sdes theory.
\newblock {\em arXiv preprint arXiv:2110.11291}, 2021.

\bibitem{chen2021stochastic}
Yongxin Chen, Tryphon~T Georgiou, and Michele Pavon.
\newblock Stochastic control liaisons: Richard sinkhorn meets gaspard monge on a schrodinger bridge.
\newblock {\em Siam Review}, 63(2):249--313, 2021.

\bibitem{dhariwal2021diffusion}
Prafulla Dhariwal and Alexander Nichol.
\newblock Diffusion models beat gans on image synthesis.
\newblock {\em Advances in neural information processing systems}, 34:8780--8794, 2021.

\bibitem{eng2021automated}
David Eng et~al.
\newblock Automated coronary calcium scoring using deep learning with multicenter external validation.
\newblock {\em NPJ digital medicine}, 4(1):88, 2021.

\bibitem{ho2020denoising}
Jonathan Ho, Ajay Jain, and Pieter Abbeel.
\newblock Denoising diffusion probabilistic models.
\newblock {\em Advances in neural information processing systems}, 33:6840--6851, 2020.

\bibitem{ho2022classifier}
Jonathan Ho and Tim Salimans.
\newblock Classifier-free diffusion guidance.
\newblock {\em arXiv preprint arXiv:2207.12598}, 2022.

\bibitem{kohl2018probabilistic}
Simon Kohl, Bernardino Romera-Paredes, Clemens Meyer, Jeffrey De~Fauw, Joseph~R Ledsam, Klaus Maier-Hein, SM~Eslami, Danilo Jimenez~Rezende, and Olaf Ronneberger.
\newblock A probabilistic u-net for segmentation of ambiguous images.
\newblock {\em Advances in neural information processing systems}, 31, 2018.

\bibitem{kurvers2016boosting}
Ralf~HJM Kurvers, Stefan~M Herzog, Ralph Hertwig, Jens Krause, Patricia~A Carney, Andy Bogart, Giuseppe Argenziano, Iris Zalaudek, and Max Wolf.
\newblock Boosting medical diagnostics by pooling independent judgments.
\newblock {\em Proceedings of the National Academy of Sciences}, 113(31):8777--8782, 2016.

\bibitem{lin1991divergence}
Jianhua Lin.
\newblock Divergence measures based on the shannon entropy.
\newblock {\em IEEE Transactions on Information theory}, 37(1):145--151, 1991.

\bibitem{liu2023i2sb}
Guan-Horng Liu, Arash Vahdat, De-An Huang, Evangelos~A Theodorou, Weili Nie, and Anima Anandkumar.
\newblock I2sb: Image-to-image schr{\"o}dinger bridge.
\newblock {\em arXiv preprint arXiv:2302.05872}, 2023.

\bibitem{pavon1991free}
Michele Pavon and Anton Wakolbinger.
\newblock On free energy, stochastic control, and schr{\"o}dinger processes.
\newblock In {\em Modeling, Estimation and Control of Systems with Uncertainty: Proceedings of a Conference held in Sopron, Hungary, September 1990}, pages 334--348. Springer, 1991.

\bibitem{rahman2023ambiguous}
Aimon Rahman, Jeya Maria~Jose Valanarasu, Ilker Hacihaliloglu, and Vishal~M Patel.
\newblock Ambiguous medical image segmentation using diffusion models.
\newblock In {\em Proceedings of the IEEE/CVF Conference on Computer Vision and Pattern Recognition}, pages 11536--11546, 2023.

\bibitem{risken1996fokker}
Hannes Risken and Hannes Risken.
\newblock {\em Fokker-planck equation}.
\newblock Springer, 1996.

\bibitem{ronneberger2015u}
Olaf Ronneberger, Philipp Fischer, and Thomas Brox.
\newblock U-net: Convolutional networks for biomedical image segmentation.
\newblock In {\em MICCAI}, pages 234--241. Springer, 2015.

\bibitem{saharia2022palette}
Chitwan Saharia et~al.
\newblock Palette: Image-to-image diffusion models.
\newblock pages 1--10, 2022.

\bibitem{selvan2020uncertainty}
Raghavendra Selvan, Frederik Faye, Jon Middleton, and Akshay Pai.
\newblock Uncertainty quantification in medical image segmentation with normalizing flows.
\newblock In {\em Machine Learning in Medical Imaging}, pages 80--90. Springer, 2020.

\bibitem{song2020score}
Yang Song, Jascha Sohl-Dickstein, Diederik~P Kingma, Abhishek Kumar, Stefano Ermon, and Ben Poole.
\newblock Score-based generative modeling through stochastic differential equations.
\newblock {\em arXiv preprint arXiv:2011.13456}, 2020.

\bibitem{szekely2013energy}
G{\'a}bor~J Sz{\'e}kely and Maria~L Rizzo.
\newblock Energy statistics: A class of statistics based on distances.
\newblock {\em Journal of statistical planning and inference}, 143(8):1249--1272, 2013.

\bibitem{vargas2021solving}
Francisco Vargas, Pierre Thodoroff, Austen Lamacraft, and Neil Lawrence.
\newblock Solving schr{\"o}dinger bridges via maximum likelihood.
\newblock {\em Entropy}, 23(9):1134, 2021.

\bibitem{vincent2011connection}
Pascal Vincent.
\newblock A connection between score matching and denoising autoencoders.
\newblock {\em Neural computation}, 23(7):1661--1674, 2011.

\bibitem{zbinden2023stochastic}
Lukas Zbinden, Lars Doorenbos, Theodoros Pissas, Adrian~Thomas Huber, Raphael Sznitman, and Pablo M{\'a}rquez-Neila.
\newblock Stochastic segmentation with conditional categorical diffusion models.
\newblock In {\em ICCV}, pages 1119--1129, 2023.

\end{thebibliography}
\endgroup

\end{document}